\pdfoutput=1

\documentclass[11pt]{article}
\usepackage{array}
\usepackage{acl}
\usepackage{times}
\usepackage{latexsym}
\usepackage[T1]{fontenc}
\usepackage[utf8]{inputenc}
\usepackage{microtype}
\usepackage{multirow}
\usepackage{booktabs}
\usepackage{amsfonts}
\usepackage{xurl}
\usepackage{subcaption}
\usepackage{blindtext}
\usepackage{graphicx}
\usepackage[export]{adjustbox}
\usepackage{wrapfig}

\title{How does fake news use a thumbnail? \\CLIP-based Multimodal Detection on the Unrepresentative News Image}

\author{Hyewon Choi\\
 Soongsil University\\\And
 Yejun Yoon\\
 Soongsil University\\\And
 Seunghyun Yoon\\
 Adobe Research\\\And
 Kunwoo Park \thanks{~~Correspondence: \texttt{kunwoo.park@ssu.ac.kr}}\\
 Soongsil University\\
 }

\begin{document}
\maketitle
\begin{abstract}
This study investigates how fake news uses a thumbnail for a news article with a focus on whether a news article’s thumbnail represents the news content correctly. A news article shared with an irrelevant thumbnail can mislead readers into having a wrong impression of the issue, especially in social media environments where users are less likely to click the link and consume the entire content. We propose to capture the degree of semantic incongruity in the multimodal relation by using the pretrained CLIP representation. From a source-level analysis, we found that fake news employs a more incongruous image to the main content than general news. Going further, we attempted to detect news articles with image-text incongruity. Evaluation experiments suggest that CLIP-based methods can successfully detect news articles in which the thumbnail is semantically irrelevant to news text. This study contributes to the research by providing a novel view on tackling online fake news and misinformation. Code and datasets are available at \url{https://github.com/ssu-humane/fake-news-thumbnail}.
\end{abstract}

\section{Introduction}

We have been suffering from the infodemic as well as the coronavirus pandemic~\cite{zarocostas2020fight}. The proliferation of fake news during the pandemic has been a significant threat to the world by inducing hate crimes against East Asians, reinforcing the wrong beliefs of anti-vaxxers, etc. Fake news is defined as ``fabricated information that mimics news media content in form but not in organizational process or intent'' \cite{doi:10.1126/science.aao2998}. Motivated by the fact that unreliable sources generate most false articles, a line of research has attempted to understand the distinct characteristics of fake news sources. A notable study is \citet{horne2017just}, which focused on textual patterns of news articles and identified that overall title structure and the use of proper nouns in titles are significant markers that differentiate fake news from general news. Similarly, from consumption and spreading patterns on social media, \citet{vosoughi2018spread} found that fake news spreads faster, deeper, and broader than general news. Other researchers showed that the reliability of news media could be predicted by various media-level features, including web traffic toward a news website~\cite{baly-etal-2018-predicting}.

\begin{figure}
    \centering
    \includegraphics[width=\linewidth]{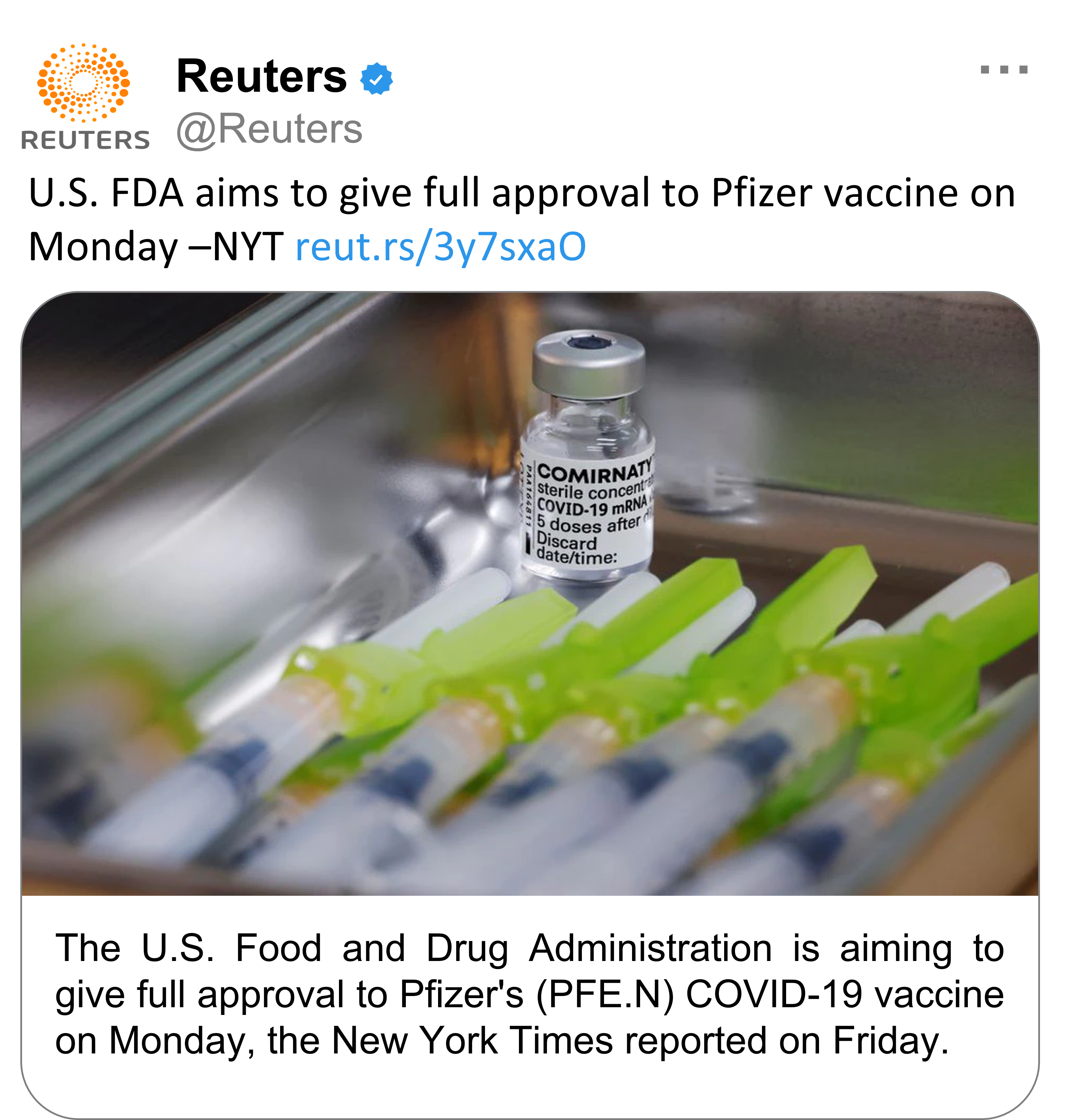}
    \caption{An example of a news article shared on Twitter. A visual summary of the article well represents the main content.}
    \label{fig:my_label}
\end{figure}

In this study, we investigate the use of images in fake news articles; in particular, we focus on a thumbnail, an image displayed as a preview to a news article. When a news article is shared on social media, its title and thumbnail image are the only visible information before a user clicks the link. Since many readers skim news without carefully checking the content~\cite{10.1145/2896377.2901462}, the visuals can mislead users into having a wrong impression if the thumbnail does not represent the news content. Fake news sources are less likely to follow the journalistic standard but tend to employ undesirable techniques such as clickbait headlines~\cite{10.1145/2823465.2823467}. Therefore, we hypothesize that unreliable sources may use a less relevant image for the thumbnail to the news text to attract clicks and promote false beliefs.

To examine the hypothesis, we propose using CLIP~\cite{radford2021learning}, a deep multimodal representation that allows representing image and text in the same embedding space. Across three datasets, we measure image-text similarity over the CLIP embedding and confirm that the fake news media tend to use the semantically less relevant photograph in news content than trustworthy sources. Going further, we test CLIP's ability to detect the incongruity between news image and text. Multi-faceted evaluation experiments highlight that the CLIP-based methods can enable article-level detection on the unrepresentative thumbnail.

We summarize the contributions of this study three-fold.

\begin{enumerate}

    \item We make a novel observation that fake news sources tend to use a less relevant news thumbnail than trustworthy media outlets.

    \item We propose a new problem for detecting misinformed news articles using semantic incongruity between news text and thumbnail.
    
    \item The paired dataset and manually annotated samples will be released for future usage.

\end{enumerate}

\section{Related Works}
\label{sec:related}

\subsection{Multimodal representation}
Researchers have explored methods that compute vector representations of multiple modalities (i.e., image and text) and align semantically similar content to the same embedding space. As examples of such attempts, building pretrained models trained with image-caption pairs shows potential as general backbone models of vision-and-language (VL) tasks~\cite{lu2019vilbert, chen2020uniter}. More recently, researchers collected large-scale image-caption data from the web and successfully trained models with a contrastive objective function. These models show robust performance in VL understanding tasks such as ``image classification'' and ``image retrieval'' even in the zero-shot setting~\cite{radford2021learning, jia2021scaling,kim2021vilt}.

As pretrained VL models can map semantically similar images and text descriptions into similar embedding spaces, they can be used to measure the quality of the image caption. Recent studies suggest a huge potential in building a better image-captioning metric using VL models~\cite {lee2020vilbertscore, lee2021umic, hessel2021clipscore}. Similarly, our study leverages the pretrained VL model to understand the relationship between news text and images. 

\subsection{Fake news detection}

Fake news detection has been actively studied in data mining and computational linguistics~\cite{10.1145/3137597.3137600}. Technically, it was tackled as a classification problem; after collecting fact-checked claims on websites such as PolitiFact\footnote{\url{https://www.politifact.com/}}, researchers trained a classification model with a wide range of features on text patterns, source characteristics, audience reactions, etc. \citet{10.5555/3061053.3061153} employed a recurrent neural network that captures patterns of contextual information of relevant posts over time. \citet{10.1145/3132847.3132877} introduced a model called CSI that incorporates the text of an article, the user response, and the source for the detection. Most recently, researchers developed a fake news detection framework that represents social contexts as a graph and learns through a graph neural network~\cite{nguyen2020fang}. This study does not aim to predict news veracity but to detect the case where the news thumbnail does not represent the main stories. 
While there have been a handful of studies tackling fake news detection using multimodal cues~\cite{8919302,8970940,9260096,10.1145/3308558.3313552}, to the best of our knowledge, no studies tackled the detection problem on incongruity between news text and image, nor investigated how fake news uses the thumbnail.

\section{Media Difference on Semantic Similarity of News Text and Image}

\subsection{Problem and hypothesis}

We aim at understanding media differences in the semantic relevance of the thumbnail picture to news text. \citet{horne2017just} suggested that fake news exhibits text patterns that are qualitatively different. Similarly, we assume that fake news may exhibit a distinct pattern in the use of news photographs:

\begin{quote}
    \emph{H. Fake news would use (semantically) a less relevant photograph to the news title for its thumbnail than general news.}
\end{quote}

We set the news title and thumbnail image, which is set as \textit{meta\_img} of the news HTML, as the target of analysis due to the following reasons. Journalism research suggests that a news title should provide a concise summary of the news article~\cite{smith1982comprehensible}, and thus we consider the title as a proxy of the news article. Among images, we use the \textit{meta\_img} because it is automatically used as a preview when being shared on social media. That is, when a news article is shared, the thumbnail picture and news title become the first content shown to the users.
Therefore, if a thumbnail does not represent the main story of a news article correctly, it could mislead readers into having a wrong impression of the target issue because social media users tend to consume news snippets without clicking the link~\cite{10.1145/2896377.2901462}. 

\subsection{Method}
\graphicspath{ {./fig/} }
\begin{figure}[ht]
    \centering
    \includegraphics[width=\linewidth]{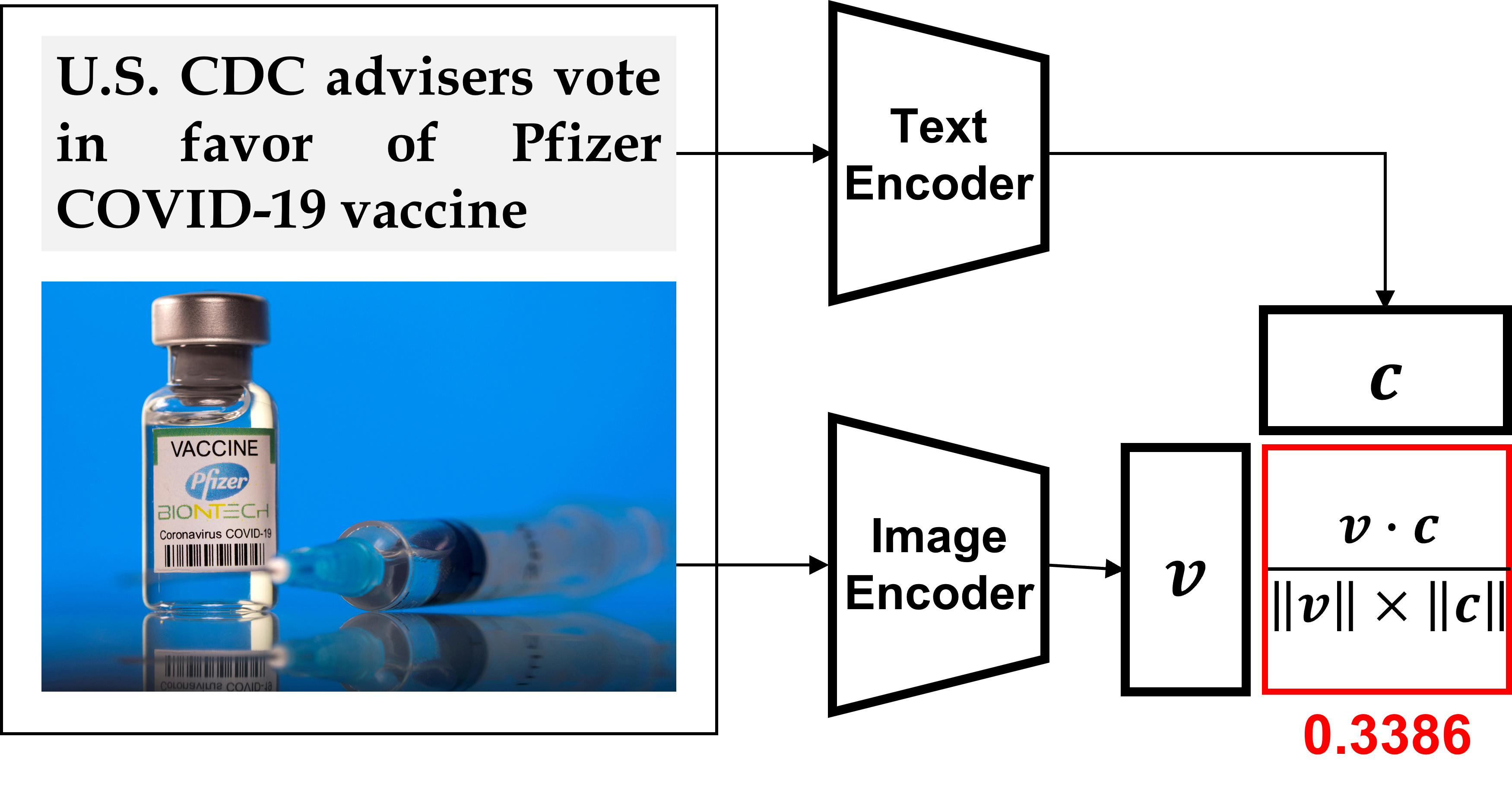}
    \caption{An illustration of CLIPScore}
    \label{fig:clipArch}
\end{figure}

To test the hypothesis, we used CLIP that represents a pair of image and text into a multimodal space~\cite{radford2021learning}, which is the state-of-the-art model in multimodal representation learning.
As shown in Figure~\ref{fig:clipArch}, we computed visual CLIP embedding $\mathbf{v}$ and textual CLIP embedding $\mathbf{c}$ of news article. Then, we measured the cosine similarity for $\mathbf{v}$ and $\mathbf{c}$ to measure their semantic relevance, also known as CLIPScore~\cite{hessel2021clipscore}\footnote{The original implementation of CLIPScore applies a parametric ReLU to the cosine similarity. We used its canonical form without the ReLU function.}.
We use the ViT-B/32~\cite{DBLP:journals/corr/abs-2010-11929} as backbone, and hence $\mathbf{c},\mathbf{v}\in \mathbb{R}^{512}$.

\subsection{Data Collection}

\begin{table}[ht]
\centering
\small
\begin{tabular}{c|ccc}
\toprule
\textbf{Type} & \multicolumn{1}{c}{\textbf{Whole}} & \multicolumn{1}{c}{\textbf{COVID}} & \multicolumn{1}{c}{\textbf{COVID-wo-faces}} \\ \midrule
General & 106,409  & 33,310 & 10,964  \\
Fake    &  3,306    & 870 & 480   \\ 
\midrule
Total   & 109,715  &  34,180  & 11,444 \\
\bottomrule
\end{tabular}
\caption{Dataset size} 
\label{tab:table1}
\end{table}

We collected news articles through the web links shared by official media accounts on social media, following a similar process proposed in a previous work~\cite{park2021present}. Our data collection pipeline consists of the following steps. 

\paragraph{Target media selection:}
To evaluate the main research hypothesis, we selected nine news outlets that run certified media accounts on Twitter as the target of analysis. Specifically, we focused on the five general news (FoxNews, New York Post, Reuters, The Guardian, Slate) and four fake news media (Activitist Post, Judicial Watch, End Time Headlines, WorldNetDaily). The target list of fake news was selected from the media sources that were labeled as \textit{red} news in a previous study~\cite{doi:10.1126/science.aau2706}, which is defined as ``spreading falsehoods that clearly reflect a flawed editorial process.'' We selected the five general news from those labeled green in the same previous work. We confirmed the general media sources considered in this study are well balanced against the political bias rating\footnote{\url{http://www.allsides.com}}. 
 
\paragraph{Tweet collection:}   
We collected tweets from January 2021 until the time of data collection (September 2021) using the Twint library\footnote{\url{https://github.com/twintproject/twint}}. We excluded tweets that do not contain URLs to their news articles.

\paragraph{News article collection:}
For each of the news URLs, we obtained the news title, body text, and URL for the thumbnail by using the newspaper3K library\footnote{\url{https://newspaper.readthedocs.io}}.
We stored the news data in JSON format and downloaded the images by the wget command. When the news data do not provide URLs for the thumbnail or we cannot download any images from the thumbnail URL, we did not include it in our data collection.

\begin{figure*}[ht]
    \centering
    \includegraphics[width=\linewidth]{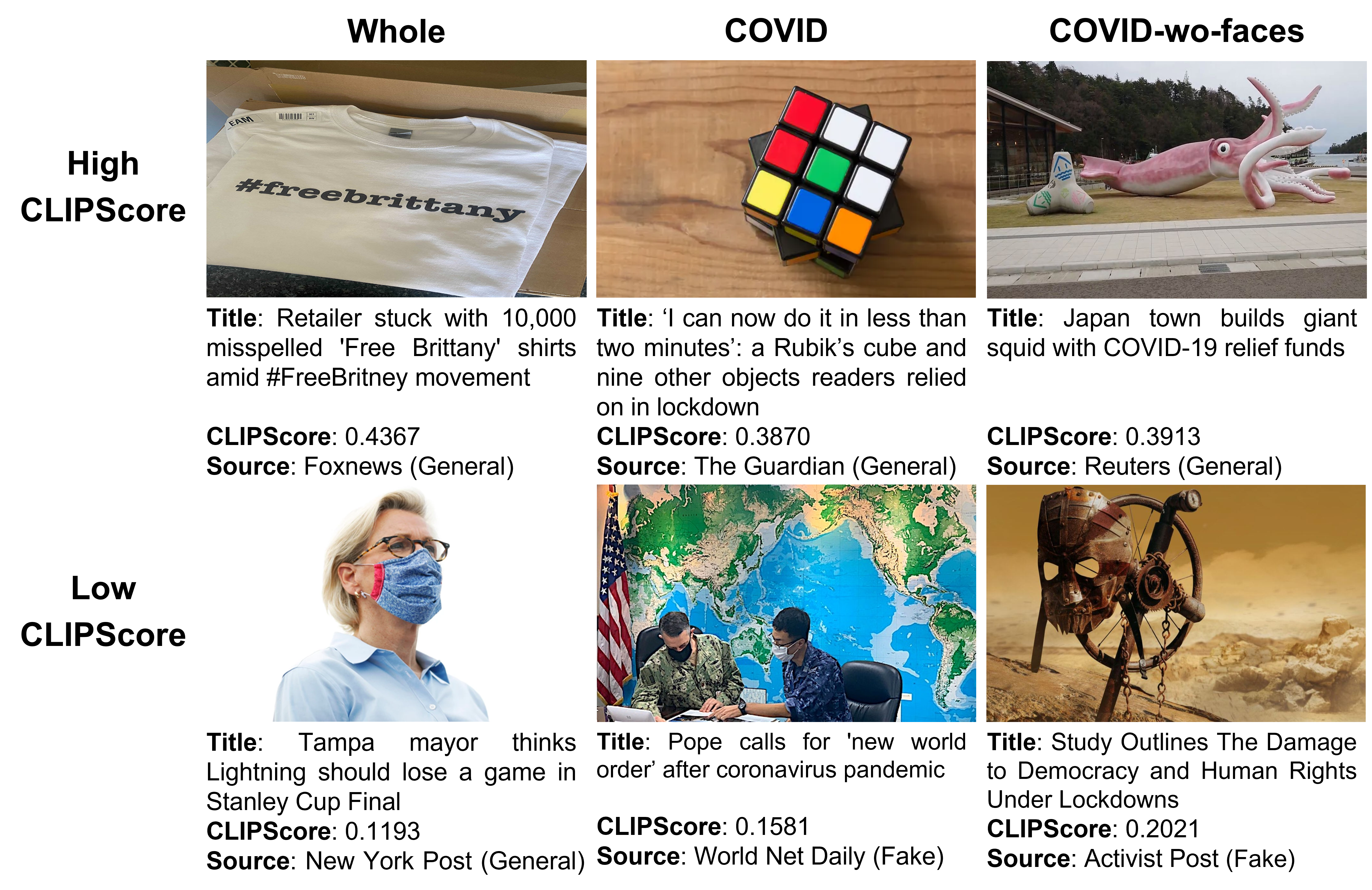}
    \caption{News examples with CLIPScore in each dataset. URLs of news articles are available in Appendix.}
    \label{fig:examples}
\end{figure*}

To see the robustness of the findings, we constructed two filtered versions of datasets for the analysis in addition to the original dataset (\textsf{Whole}). First, we limited the scope of the news topic to COVID-19 by selecting news articles containing at least one of the COVID-19 related keywords: coronavirus, corona, covid-19, corona virus, covid, covid19, sars-cov-2, pandemic, chinese virus, chinesevirus, and corona. The COVID-19 issue has been covered extensively during the period of CLIP training, and thus we assumed the CLIP embedding could understand the COVID-19 context better than random events. We call the COVID-19 filtered dataset \textsf{COVID}. Next, to minimize the number of false negatives (i.e., the model considers a relevant pair irrelevant), we further filtered out news articles in which the thumbnail picture contains faces from the COVID dataset (\textsf{COVID-wo-faces}. In a preliminary analysis, we found that CLIP is not good at matching a person's name in text and their appearance in an image, especially when they are not famous (e.g., the example in the bottom left of Figure~\ref{fig:examples} and Figure~\ref{fig:appendix1}.)). We detected images with a face by the face detection model of the Google Cloud Vision\footnote{\url{https://tinyurl.com/ydfu2js3}}. Table~\ref{tab:table1} presents the size of three datasets that covers news articles from January to August 2021. We expect that the data leakage issue is minimal because our dataset period is less likely to overlap with the dataset used for training CLIP\footnote{CLIP paper was released on Feb 26th, 2021, which does not explicitly mention the period of the training dataset.}.

\begin{figure*}[t]
     \centering
     \begin{subfigure}{.34\linewidth}
         \includegraphics[width=\linewidth, height=3.3cm]{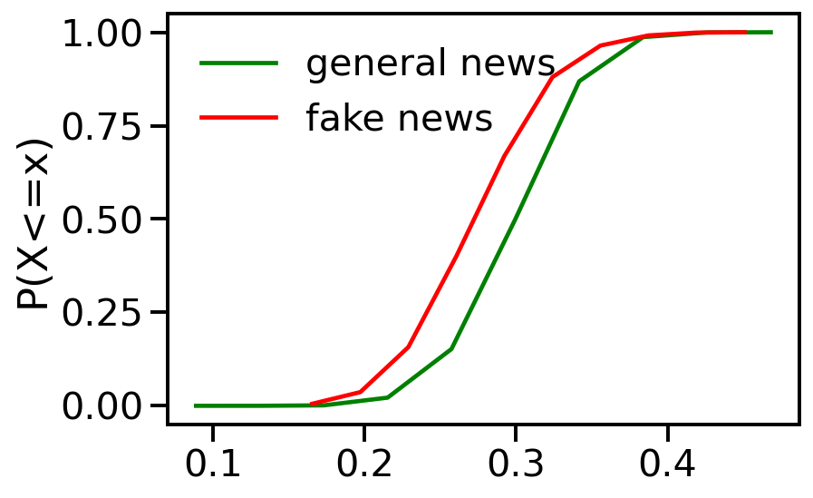}
          \caption{Whole (0.596$^{***}$)}
         \label{fig:whole}
     \end{subfigure}
     \hfill
     \begin{subfigure}{.3\linewidth}
         \includegraphics[width=\linewidth, height=3.3cm]{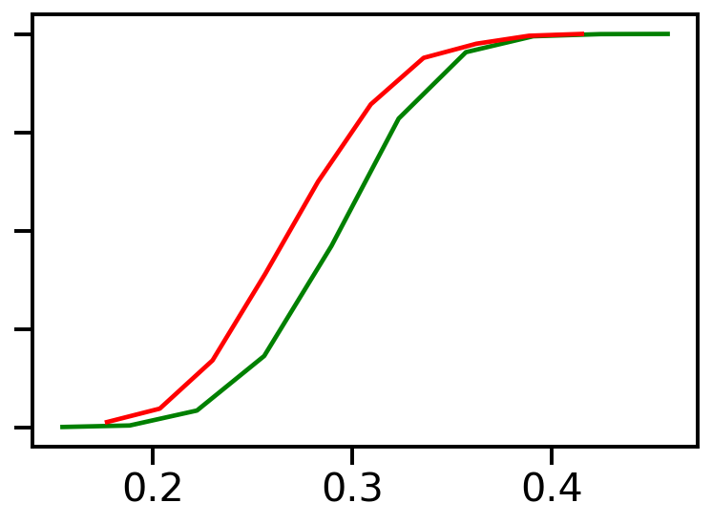}
         \caption{COVID (0.545$^{***}$)}
         \label{fig:covid}
     \end{subfigure}
     \hfill
     \begin{subfigure}{.3\linewidth}
         \includegraphics[width=\linewidth, height=3.3cm]{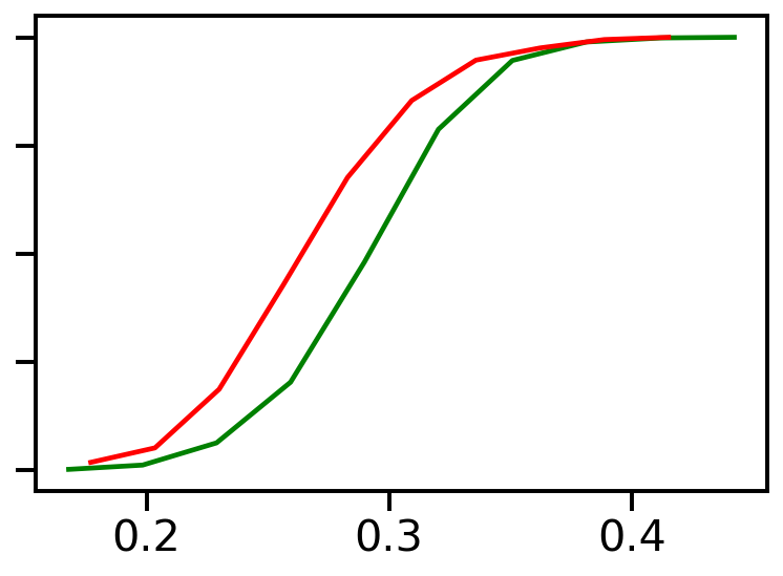}
         \caption{COVID-wo-faces (0.594$^{***}$)}
         \label{fig:covidface}
     \end{subfigure}
        \caption{CDFs of the CLIPScore measured for each dataset. Values within the parenthesis indicate Cohen's d corresponding to the difference of CLIPScore between general and fake news ($^{***}$: \emph{p}$<$0.001 by the t-test).}
        \label{fig:cdfs}
\end{figure*}

\subsection{Results}

Figure~\ref{fig:examples} presents the title-image pairs with the CLIPScore values. The three examples in the top row present the pairs with a high CLIPScore, which were sampled from the top-500 news articles in terms of CLIPScore. The bottom three examples were randomly selected from the bottom-500 examples in terms of CLIPScore. The high-score examples demonstrate the capability of CLIP in understanding a written text and the appearance of a visual object. On the other hand, the three examples at the bottom demonstrate two scenarios where a low CLIPScore can represent. First, the New York Post example from the whole dataset suggests that the CLIP encoder has difficulty recognizing a person's appearance in an image, a name in a text, or both. Second, the low-score examples for the COVID and COVID-wo-faces datasets represent the cases where a thumbnail does not represent the news text, suggesting the potential of CLIPScore for capturing news articles with an unrepresentative thumbnail. Therefore, we used CLIPScore for understanding the media difference between fake news and trustworthy media in terms of semantic relevance between news title and thumbnail across the three datasets. The observations from the filtered datasets can function as a robustness check.

Figure~\ref{fig:cdfs} presents the difference of the semantic relevance of news title and thumbnail between fake and general news, measured by CLIPScore. We conducted the t-test to evaluate the statistical significance of a difference and calculated the Cohen's d for its effect size. The x-axis presents the CLIPScore threshold, and the y-axis presents the probability that the CLIPScore takes a value less than or equal to the threshold from the distribution. Results indicate that fake news tends to have a lower CLIPScore than general news with a statistical significance across the three datasets. The corresponding effect size is 0.596, 0.545, and 0.594 for the Whole, COVID, and COVID-wo-faces dataset, respectively. The values are considered medium effect sizes, which suggests that fake news tends to use a thumbnail picture that is semantically less similar to the news title than general news and therefore supports the main hypothesis in \S 3.1.

\section{Detection of News Articles with the Incongruous Image}

\subsection{Motivation}
As we observed in the previous section, Fake news media tend to use a photograph that is semantically less relevant to the news text than general news. Motivated by the observation, we turned to a detection problem aiming at identifying news articles with the incongruous thumbnail among articles shared by fake news outlets. We focused on the scope of detection of fake news media because the potential negative impact of image-text incongruity can be worse when used to promote false claims. Also, previous research suggested visuals can give a more significant impression to readers than textual signals~\cite{seo2020meta}.

Formally, we define the problem as a classification task using image-text multimodal data: given a pair of news text $T$ and image $I$, we aim at predicting the binary incongruity label $L$ on whether $I$ is semantically (in-)congruent with $T$.

\subsection{Data generation}

\begin{figure}[ht]
    \centering
    \includegraphics[width=\linewidth]{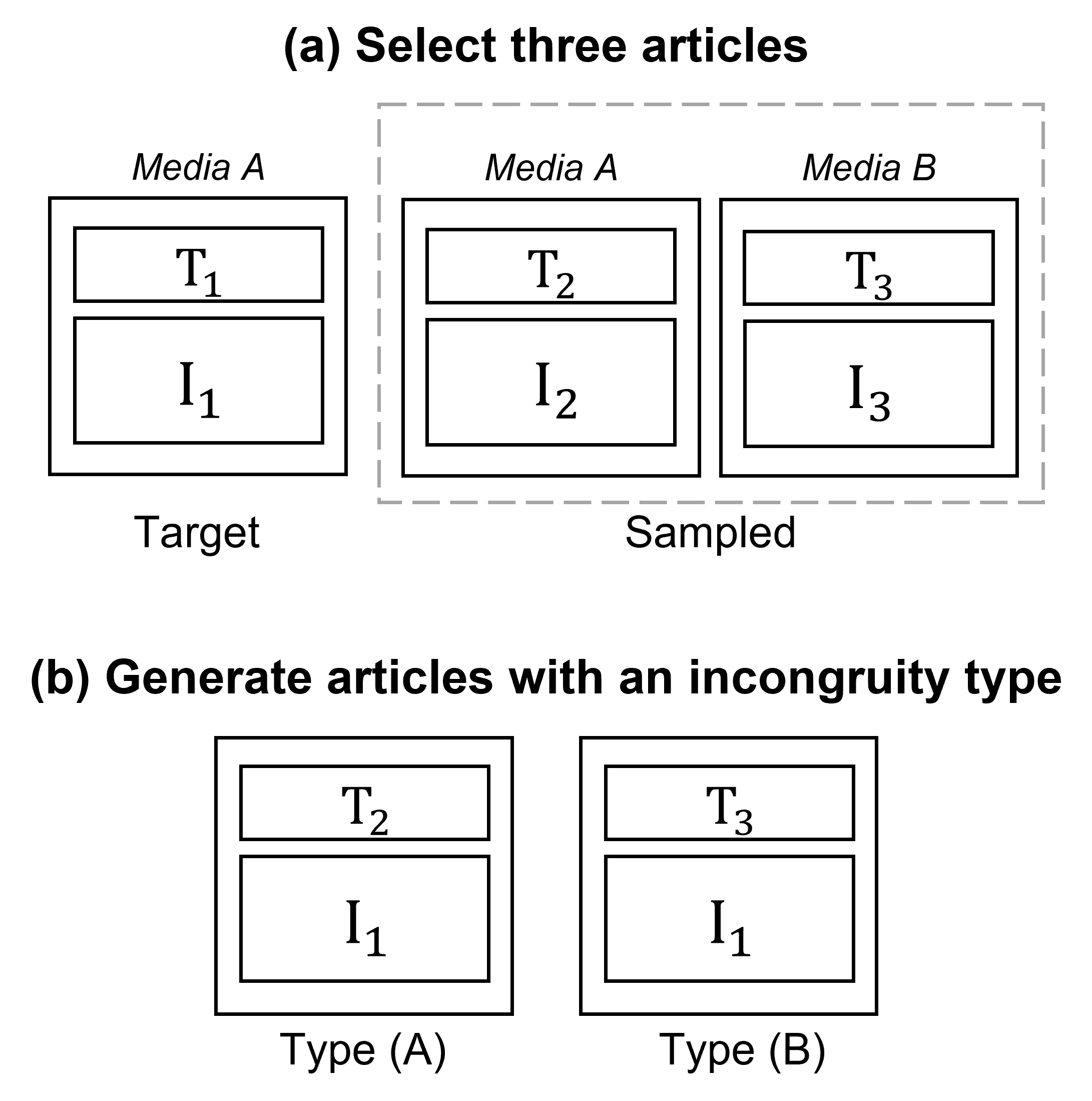}
    \caption{An illustration of data generation process ($T$: news title, $I$: thumbnail image).}
    \label{fig:datagen1}
\end{figure}

A significant challenge in implementing a classification model for the target task is the lack of a dataset. While we have more than 20k image-text pairs, they are unlabeled, and it is costly to annotate the incongruity label for all the pairs manually. Therefore, inspired by a previous study~\cite{yoon2019detecting}, we utilized an alternative method that generates a pair of $I$ and $H$ with the incongruity label $L$ automatically. The data generation method is language-agnostic, such that it can be easily extended to any other language as long as one can construct a pool of trustworthy news articles.

Figure~\ref{fig:datagen1} demonstrates the data generation process. At first, among the news articles generated by trustworthy news sources in the COVID-wo-faces dataset, we selected the top 75\% of the image-text pairs in terms of CLIPScore to be congruent samples. As a result, we obtained 8223 target samples. We manually inspected the bottom-100 samples and confirmed that the image represents the news content well. To be used for generating train/validation/test datasets in the next step, we divided the 8223 pairs into three pools: 6575, 824, and 824, respectively.

The next step is to generate news articles with the incongruity between news title and thumbnail. As shown in Figure~\ref{fig:datagen1}(a), for each pair in the congruent dataset, we randomly sampled two different pairs, one from the same media and another from one of the other outlets. We called the two pairs \textit{sampled}. Then, as in Figure~\ref{fig:datagen1}(b), we automatically generated samples with the incongruity by linking the image of the target article ($I_1$) to the title of the sampled articles ($T_2$, $T_3$). That is, the class ratio is 2:1 in the dataset. We applied the generation process to each pool separately, and therefore there are no overlapped articles between one dataset to another.

In total, we obtained 8223 congruent and 16446 incongruent pairs, and there are 19725, 2472, and 2472 samples for train/validation/test, respectively.  

\begin{figure}[t]
    \centering
    \includegraphics[width=\linewidth]{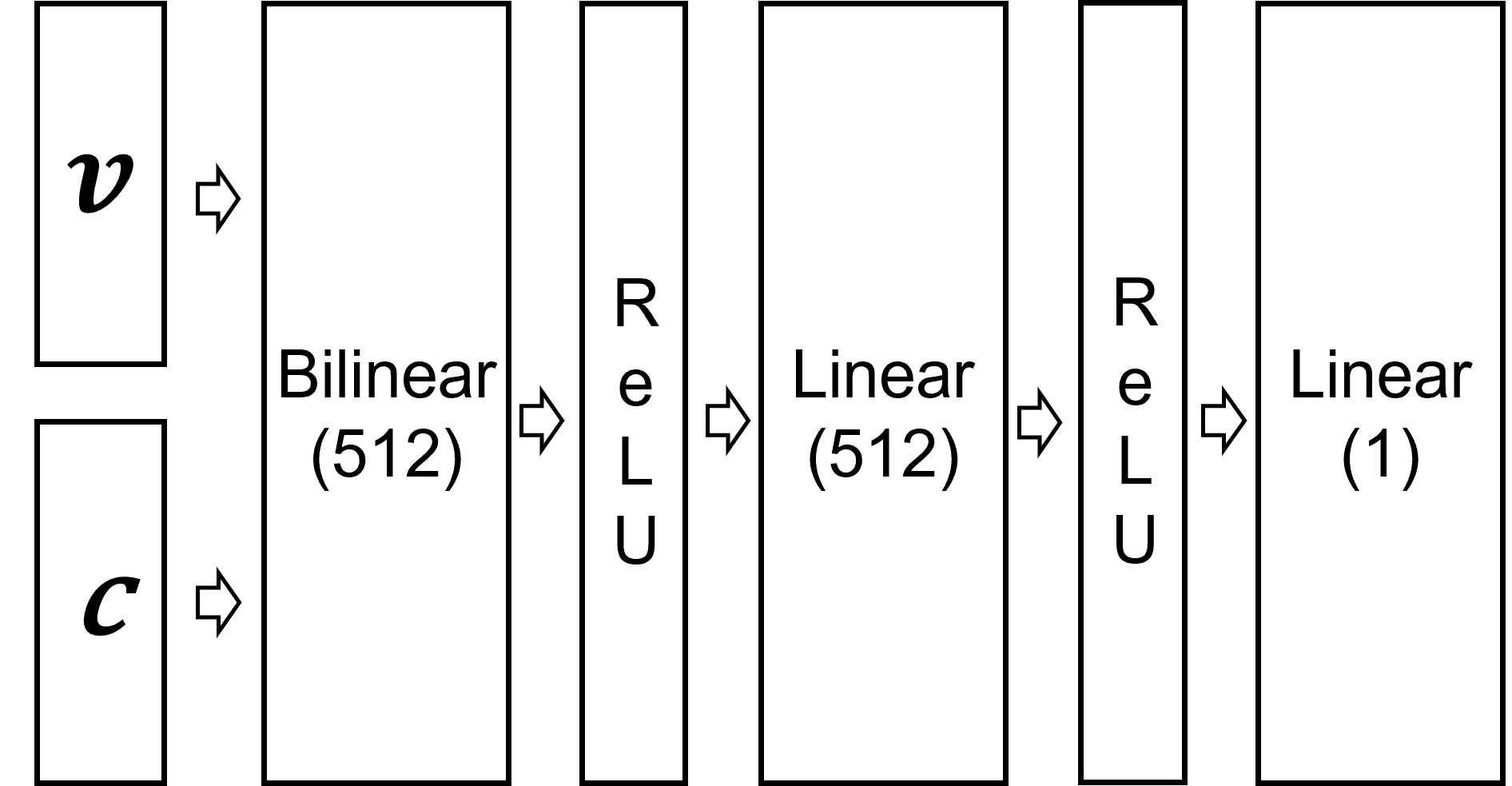}
    \caption{CLIP-classifier’s model architecture. The value within parenthesis indicates the output dimension size.}
    \label{fig:CLIP-classifier}
\end{figure}

\subsection{Experimental Results}
We used a machine equipped with the AMD Ryzen Threadripper Pro 3975WX CPU and two Nvidia RTX A6000 GPUs for the experiments. We evaluated three different methods for detecting image-text incongruity among fake news articles.

\begin{itemize}
    \item ViLT (zero-shot): As a baseline model, we employed a recent vision-and-language pretrained model, ViLT \cite{kim2021vilt}, which was fine-tuned on the MS COCO dataset. Using the cosine similarity between image and text vectors, we implemented a simple threshold-based classifier; If a similarity value is above the threshold, the model predicts the text well represents the image. Otherwise, a pair is considered unmatched. We obtained the decision threshold by a class-wise unweighted average for the similarity scores measured on all samples in the validation set. The obtained threshold was also used for test set inference.

    \item CLIPScore (zero-shot): Using the pretrained CLIP model, we computed the CLIPScore for each news title and thumbnail pair for implementing a threshold-based classifier. The decision threshold was obtained following the same procedure used for ViLT (zero-shot).
 
    \item CLIP-classifier: Figure~\ref{fig:CLIP-classifier} shows the neural architecture of the proposed model. CLIP-classifier takes as input \textbf{\textit{c}} (text embedding) and \textbf{\textit{v}} (visual embedding) from CLIP's text and visual encoder, respectively, and classifies the pair as `congruent' (well-matched) or `incongruent' (not-well-matched). The model was trained to minimize the binary cross-entropy loss by the AdamW optimizer (at a learning rate of 0.001) with a batch size of 128. We did not update the CLIP backbone during training. We used gradient clipping with a threshold of 1.0 and early stopping.
\end{itemize}

\begin{table}[t]
\small
\resizebox{1.00\columnwidth}{!}{%
\begin{tabular}{ccccccc}
\toprule
\multirow{2}{*}{Model} & \multicolumn{2}{c}{Validation} & \multicolumn{2}{c}{Test} \\
\cmidrule{2-5}
                    & ACC.    & AUROC   & ACC.  & AUROC \\
\midrule
ViLT (zero-shot)                & 0.646 & 0.720 & 0.601 & 0.700 \\
CLIPScore (zero-shot)           & 0.942   & 0.985   & 0.934 & 0.984    \\
CLIP-classifier      & 0.920 & 0.977 & 0.927 & 0.975 \\
\bottomrule
\end{tabular}%
}
\caption{Evaluation on the generated set.}
\label{tab:evaluation}
\end{table}

Table~\ref{tab:evaluation} presents the evaluation results of the three models. The two CLIP-based models outperformed ViLT (zero-shot) with a large margin. These observations suggest that the CLIP pretrained model is more generalizable than the ViLT model, and hence it is more suitable for the detection of fake news articles that use an unrepresentative thumbnail.

\begin{figure}[t]
    \centering
    \includegraphics[width=\linewidth]{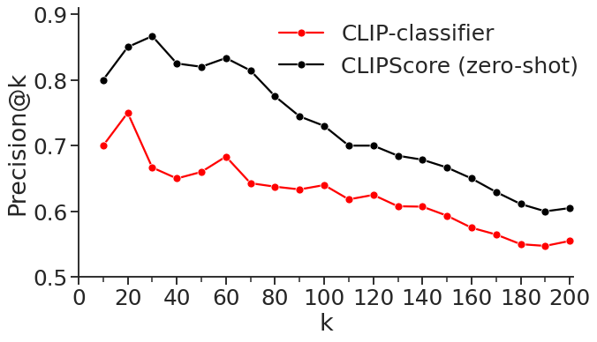}
    \caption{Manual evaluation results on the top-k fake news articles by CLIPScore and CLIP-classifier's predictions score}
    \label{fig:top100}
\end{figure}

To test the ability of CLIP in real-world detection, we conducted additional experiments with human annotations. We supposed a situation where it is required to detect fake news articles using the incongruous thumbnail. Hence, we inferred prediction scores for the fake news samples in the COVID-wo-faces dataset by CLIPScore and CLIP-classifier, respectively. Then, we manually inspected the top-200 examples of each model in terms of the prediction score to test whether the models correctly predict the samples of an unrepresentative thumbnail. We considered the incongruous label as the positive label; Hence, a higher prediction score indicates a model predicts a given pair having the incongruity between news title and thumbnail picture with higher confidence. For consistency, we used $(1-similarity)$ for the prediction score of CLIPScore.

Figure~\ref{fig:top100} shows the top-k precision of each model's prediction on the fake news articles. The x-axis represents the number of evaluated articles after being sorted by a model's prediction score. The y-axis shows the precision of the top-k articles evaluated by humans. Two authors participated in the manual annotation process and obtained a complete inter-annotator agreement after several iterations. They examined a total of 259 news-thumbnail pairs on whether the image represents the news content. We release the paired dataset with manual annotation for broader usage on the github repository.

Results show that CLIPScore outperformed CLIP-classifier, especially for the highly-ranked examples. The model achieved a precision of 0.8 for k=10, 0.85 for k=20, and 0.87 for k=30; its performance gap against CLIP-classifier is around 0.1. The gap decreased as more examples were evaluated; the precision difference is 0.05 for k=200. The observation highlights the representation power of the CLIP backbone and implies that the two CLIP-based methods could be incorporated for more effective detection in practice. 

\section{Limitation and Future Direction}

This study bears several limitations. First, the findings were observed from the dataset of nine news media. Even though they are well-balanced against political bias and trustworthiness, the findings could not represent general patterns and thus should be carefully interpreted. Future studies could examine the hypothesis using more extensive data. Second, since this study employs CLIP as a backbone, our results are subject to unknown biases which CLIP might learn from training. Future studies could adopt pretraining tasks to mitigate the issues. Third, we focused on news titles as a proxy of news content. The method could be invalid for some cases where the news title is incongruent with the main text~\cite{yoon2019detecting}. Future studies could develop a method that exploits body text as a reference, which contains more fruitful information yet is more challenging to be analyzed.

\section{Conclusion}

This paper examined the usage of news thumbnails and asked whether fake news sources exhibit distinct patterns. By applying CLIP to the pair of news title and image, we identified the difference between fake news and trustworthy media sources in the image-text similarity: Fake news tends to use a less similar thumbnail picture to the news text than general news. Next, we tackled the article-level detection problem that targets fake news articles in which the thumbnail picture does not represent the news content. To the end, we generated a paired dataset of 24,669 image-text pairs, each image of which is semantically (in-)congruent to the text. Evaluation experiments showed that CLIP-based models could detect news articles with an unrepresentative thumbnail with high accuracy. These observations highlight the potential of CLIP for identifying these misinformed articles in the real world. To the best of our knowledge, this is one of the initial attempts to understand fake news characteristics in the use of thumbnail and focus on its semantic representativeness to news content. We hope our methodology and dataset can not only make an impact on the ongoing efforts to curtail fake news dissemination, but also contribute to broader research communities on vision and language. 

\section*{Acknowledgements}
H. Choi and Y. Yoon equally contributed to this work. This work was supported by the Basic Science Research Program through the National Research Foundation of Korea (NRF) funded by the Ministry of Science and ICT (No. NRF-2021R1F1A1062691).

\bibliography{custom}
\bibliographystyle{acl_natbib}

\appendix

\renewcommand{\thefigure}{A\arabic{figure}}

\setcounter{figure}{0}
\setcounter{table}{0}

\section{Appendix}
\label{sec:appendix}

\begin{figure}[ht]
    \centering
    \includegraphics[width=\linewidth]{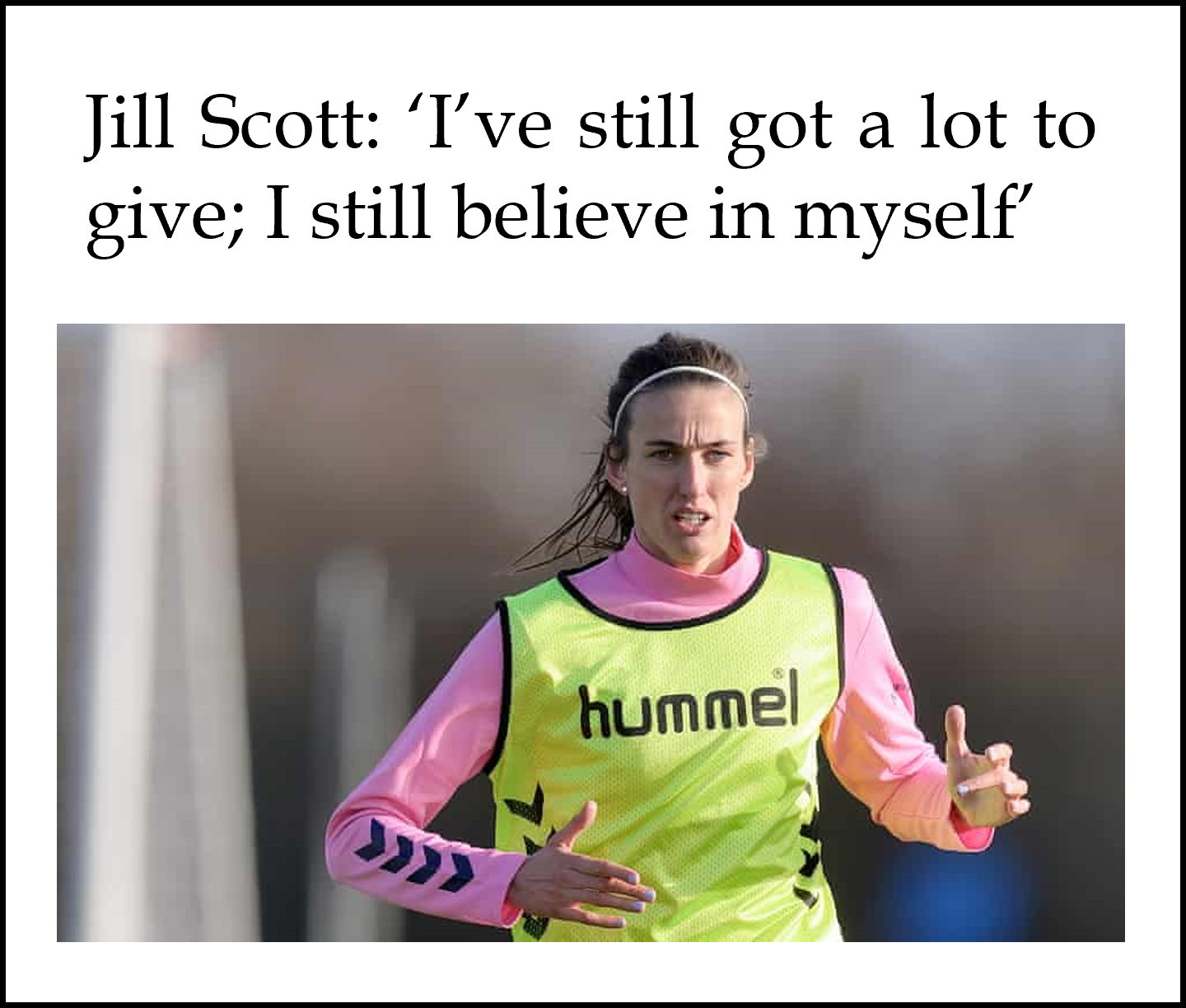}
    \caption{An example of news article that CLIP has difficulty at matching a person's name and face (CLIPScore: 0.04694, URL: https://tinyurl.com/y4y89b3x).}
    \label{fig:appendix1}
\end{figure}

\setcounter{table}{0}
\renewcommand{\thetable}{A\arabic{table}}
\label{sec:appendix2}

\begin{table}[ht]
\resizebox{1.00\linewidth}{!}{
\begin{tabular}{c|c|c|l}
\multicolumn{1}{l|}{}                    & \multicolumn{1}{c|}{\textbf{CLIPScore}} & \multicolumn{1}{c|}{\textbf{Source}} & \multicolumn{1}{c}{\textbf{URL}} \\ \hline
\multirow{2}{*}{\textbf{Whole}}  & High & Foxnews & https://tinyurl.com/ydrc32kl     \\
                                 & Low & New York Post  & https://tinyurl.com/y7794djr     \\ \hline
\multirow{2}{*}{\textbf{COVID}}  & High & The Guardian & https://tinyurl.com/y8r7o2b7     \\
                                 & Low & World Net Daily & https://tinyurl.com/yalznxnn     \\ \hline
\multirow{2}{*}{\textbf{\begin{tabular}[c]{@{}c@{}}COVID-\\ wo-faces\end{tabular}}} & High   & Reuters  & https://tinyurl.com/ydozsybd     
\\ & Low   & Activist Post  & https://tinyurl.com/ycjkbell  \\ \hline 

\end{tabular}
}
    \caption{URLs for news articles in Figure~\ref{fig:examples}}
\end{table}

\end{document}